\documentclass[10pt,twocolumn,letterpaper]{article}

\usepackage{cvpr}              

\usepackage[table,dvipsnames]{xcolor}
\usepackage{colortbl} 
\definecolor{TableDarkGreen}{RGB}{182,215,168}
\definecolor{TableLightGreen}{RGB}{207,234,215}
\definecolor{TableYellow}{RGB}{255,242,204}
\definecolor{TableRed}{RGB}{244,204,204}

\usepackage{array}
\usepackage{multirow, makecell}
\usepackage{arydshln}
\usepackage{hhline}
\usepackage[most]{tcolorbox} 
\usepackage[export]{adjustbox}

\newcommand{\TCBTableDarkGreen}[1]{\tcbox[on line,boxsep=0pt,left=0pt,right=0pt,top=0pt,bottom=0pt,colframe=white,colback=TableDarkGreen,highlight math style={enhanced},extrude by=0.75pt]{#1}}

\usepackage{soul}

\definecolor{cvprblue}{rgb}{0.21,0.49,0.74}
\usepackage[pagebackref,breaklinks,colorlinks,allcolors=cvprblue]{hyperref}

\usepackage[ruled,vlined,linesnumbered]{algorithm2e}

\def\paperID{12113} 
\def\confName{CVPR}
\def\confYear{2026}

\title{Learning Scene Coordinate Reconstruction from Unposed Images \\ via Pose Graph Optimization}

\author{Tze Ho Elden Tse$^{1,2}$,\quad Jizong Peng$^1$,\quad Angela Yao$^2$\\
$^1$dConstruct Robotics \quad $^2$National University of Singapore\\
\small\url{https://eldentse.github.io/visual_localization_pgo_public/}
}

\begin{document}
\maketitle
\begin{abstract}
Learning-based structure-from-motion methods such as ACE-Zero have demonstrated strong performance in estimating camera poses and scene coordinates from unordered image collections without requiring ground truth supervision. However, the lack of global and multi-view consistency constraints in ACE-Zero can lead to pose drift and misalignment, particularly in complex or ambiguous scenes. In this work, we propose a hybrid framework that integrates pose graph optimization (PGO) into ACE-Zero to refine camera poses and suppress incorrect refinements. We construct pose graphs directly from ACE-Zero outputs by extracting relative pose constraints from predicted scene coordinates. Furthermore, we introduce an uncertainty-aware optimization strategy by estimating confidence scores using geometric priors, including epipolar and optical flow consistencies across views. Our approach improves the robustness and accuracy of pose estimation, demonstrating that global geometric reasoning can effectively complement learning-based inference in structure-from-motion.
\end{abstract}    
\section{Introduction}
\label{sec:intro}

Structure-from-Motion (SfM) and visual localization are fundamental problems in computer vision and robotics, enabling applications such as 3D reconstruction, autonomous navigation, and augmented reality~\cite{peng2025constrained,xu2024das3r,sun2025visual,feng2023diffpose,tse2019no}. Recent advances~\cite{wang2025vggt,brachmann2024acezero,wang2024vggsfm} have introduced learning-based approaches that bypass traditional feature matching and bundle adjustment.
Among these, ACE-Zero~\cite{brachmann2024acezero} has emerged as a promising visual localization framework that estimates camera poses and scene coordinates directly from unordered image collections, without requiring ground truth or explicit geometric supervision.

\begin{figure}[t]
\centering
\includegraphics[width=1\linewidth]{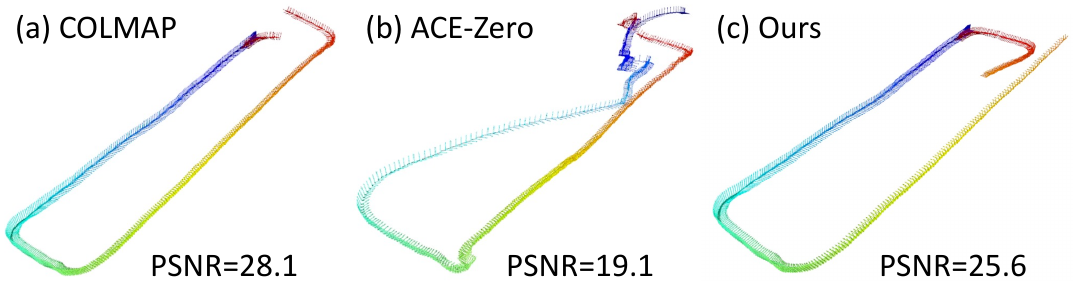}
\caption{Despite being initialized from (a) COLMAP generated camera poses, local refinements in (b) often introduce global inconsistencies. In contrast, (c) our hybrid-framework, which integrates pose graph optimization into ACE-Zero, enforces multi-view consistency and suppress incorrect refinements. 
}
\vspace{-0.2cm}
\label{fig:teaser}
\end{figure}

As shown in Fig.~\ref{fig:teaser}, ACE-Zero's reliance on local neural refinement can lead to locally optimal but globally inconsistent poses, particularly in large-scale or ambiguous scenes. The absence of global consistency constraints makes the framework susceptible to drift and misalignment, especially when the initial pose estimates are noisy or when scene coordinate predictions are uncertain. 

To address these limitations, we propose a hybrid framework that integrates pose graph optimization (PGO) into the ACE-Zero pipeline. PGO is a well-established technique in SLAM and SfM that enforces global consistency by optimizing camera poses over a graph of relative constraints~\cite{carlone2015initialization,carlone2014fast}. Unlike purely local refinement, PGO leverages pairwise pose relationships across the entire scene, making it well-suited for correcting accumulated drift and resolving ambiguities. 
Compared to alternatives such as global bundle adjustment or learning-based refinement, PGO is generally more scalable due to its reduced parameter space as it optimizes only camera poses rather than both poses and 3D structure. Moreover, its graph-based formulation offers clearer interpretability by explicitly modeling pairwise pose constraints, which can be visualized and debugged more easily than opaque learned representations.


However, integrating PGO into ACE-Zero is non-trivial. ACE-Zero does not produce explicit pairwise pose constraints or co-visibility graphs, which are typically required for graph construction. Moreover, the scene coordinate predictions are inherently noisy and lack uncertainty estimates, making naive optimization prone to over-fitting or instability. To overcome these challenges, we introduce a geometry-driven approach to derive reliable relative pose constraints from ACE-Zero outputs and construct a robust pose graph for optimization. Specifically, we identify image pairs with consistent scene coordinate predictions and use them to construct the graph edges for PGO.

Furthermore, we extend our framework with uncertainty-aware optimization, which incorporates confidence estimates into the graph refinement process. This allows the optimizer to weigh constraints based on their reliability, suppressing the influence of uncertain predictions and improving robustness. While this requires additional computation to estimate and propagate uncertainty, it significantly enhances the quality of the final pose estimates, especially in scenes with occlusions, repetitive structures, or low texture. These conditions often lead to ambiguous or unreliable 2D-3D correspondences, which can severely degrade pose accuracy. By integrating uncertainty modeling, our approach reduces the impact of such ambiguities and improves robustness in these challenging environments.


This work bridges the gap between learning-based and optimization-based SfM, showing that global geometric reasoning can complement neural inference to produce more reliable and consistent camera pose estimates. 

We summarize our main contributions as follows:
\begin{itemize}
    \item A hybrid framework that integrates pose graph optimization into ACE-Zero to enforce multi-view and global consistency.
    \item A method for constructing pose graphs from ACE-Zero outputs without explicit geometric supervision.
    \item An uncertainty-aware optimization strategy that improves robustness by accounting for prediction confidence.
    \item Our framework achieves more reliable camera poses estimates while requiring on average half the number of iterations compared to ACE-Zero alone.
\end{itemize}


\section{Related Work}
\label{sec:related}


\noindent\textbf{Traditional SfM} pipelines~\cite{hartley2003multiple,oliensis2000critique,ozyecsil2017survey,schoenberger2016sfm} rely on feature matching, triangulation, and bundle adjustment to estimate camera poses and reconstruct 3D scenes. While effective, these methods are sensitive to textureless regions, occlusions, and large viewpoint changes.

\noindent\textbf{Learning-based SfM} approaches~\cite{wang2025vggt,brachmann2024acezero,wang2024vggsfm} aim to overcome these limitations by directly regressing scene coordinates or camera poses from the images. Notably, ACE-Zero~\cite{brachmann2024acezero} estimates camera parameters from unordered image collections using a neural scene coordinate regression model. By predicting 3D coordinates directly from images and using reprojection-based losses instead of ground-truth supervision, ACE-Zero enables self-supervised learning without relying on explicit feature matching or pose priors.
However, we observe that ACE-Zero can suffer from pose drift and misalignment in complex scenes. We hypothesize that this is due to the absence of global consistency constraints, as its refinement is purely local and does not leverage sequential or pairwise structure. To address this, we propose a hybrid framework that integrates pose graph optimization to enforce multi-view consistency and improve robustness.

Other learning-based SfM methods such as DeepSFM~\cite{wei2020deepsfm} and GraphSfM~\cite{brynte2024gasfm} incorporate neural architectures to replace traditional submodules like triangulation and pose averaging. These methods often assume structured inputs (\ie image sequences or reliable initialization) which inherently provide stronger geometric priors and facilitate global consistency. In contrast, ACE-Zero operates on unordered image sets without pose supervision, making it more susceptible to global misalignment.


\noindent\textbf{Pose Graph Optimization (PGO)} is a core technique in SLAM and global SfM.  It refines camera poses by minimizing the inconsistency between relative transformations. Each node in the graph represents a camera pose, and each edge encodes a relative constraint derived from image pairs. Optimization is typically performed over the Lie group $SE(3)$ or its variants, such as Sim(3) when scale ambiguity is present~\cite{barath2021posegraph, sim3posegraph2025}. PGO has been widely adopted in SLAM systems like ORB-SLAM~\cite{mur2015orb} and GTSAM-based pipelines~\cite{dellaert2012factor}, where it helps enforce loop closure and correct accumulated drift.
Recent works have explored efficient initialization~\cite{barath2021posegraph}, robust estimators, and graph-theoretic principles to improve scalability and convergence. However, integrating PGO into learning-based SfM frameworks remains underexplored, especially in settings without ground truth or sequential data.


\noindent\textbf{Uncertainty-aware optimization} incorporates uncertainty into the pose estimation for both object-centric scenarios (\eg, single-instance localization) and scene-level reconstruction (\eg, multi-view SfM). Methods like UA-Pose~\cite{li2025uapose} and UnPose~\cite{jiang2025unpose} leverage uncertainty estimates to guide optimization and improve robustness to noisy inputs. These approaches typically derive uncertainty from neural networks or diffusion models.
Uncertainty-aware formulations of PGO have been explored in domains such as robotics and object pose estimation, where edge weights are adjusted based on measurement confidence~\cite{merrill2022symmetry,li2021pogo}. However, their application to learning-based SfM remains under-explored, primarily due to the lack of explicit uncertainty estimates and structured pairwise constraints in neural models like ACE-Zero.

\noindent\textbf{Visual localization} 
aims to estimate the camera pose of a query image within a previously reconstructed scene. Classical approaches~\cite{sattler2011fast,sattler2012improving,sattler2016efficient} rely on matching 2D features from the query to 3D points in the scene, followed by pose estimation via PnP. For large-scale environments, retrieval-based methods~\cite{sarlin2019coarse,sattler2017large} improve efficiency by narrowing down relevant database images. 
Learning-based methods offer alternatives by encoding scene geometry directly into neural networks. PoseNet~\cite{kendall2015posenet} regresses absolute poses directly from images, but struggles with generalization beyond training poses~\cite{sattler2019understanding}. Relative pose regression~\cite{balntas2018relocnet,turkoglu2021visual} and scene coordinate regression~\cite{brachmann2024acezero,brachmann2017dsac,cavallari2017fly,bruns2025aceg,bian2025scrpriors} improve flexibility by estimating geometric relationships or 3D coordinates, enabling pose estimation via PnP and RANSAC. These methods are typically trained in a supervised manner, relying on ground-truth poses or 3D annotations. In contrast, ACE-Zero~\cite{brachmann2024acezero} trains a visual localizer from unposed images. Our work builds on ACE-Zero by extending it with global consistency via pose graph optimization and uncertainty-aware refinement.

\noindent\textbf{Hybrid approaches} that combine learning-based inference with optimization-based refinement are emerging as a powerful paradigm in visual localization. HERO-SLAM~\cite{xin2024heroslam} and DeepSFM~\cite{wei2020deepsfm} demonstrate that neural representations can be enhanced by geometric reasoning, leading to improved accuracy and robustness. These methods highlight the complementary strengths of data-driven and model-based techniques. In this work, we build on this insight by integrating pose graph optimization into ACE-Zero, using learned scene coordinates to construct relative constraints and geometric priors to estimate confidence. This hybrid framework bridges the gap between implicit neural representations and explicit geometric consistency.

\section{Methodology}

\subsection{Preliminary}

\noindent\textbf{Pose graph optimization (PGO).}  PGO is a widely used technique in structure-from-motion (SfM) and simultaneous localization and mapping (SLAM) to refine camera poses by minimizing inconsistencies across pairwise relative transformations.
A pose graph is defined as a set of nodes and edges:

\begin{itemize}
    \item \textbf{Nodes:} Each node $i$ represents a camera pose $\mathbf{T}_i = [\mathbf{R}_i | \mathbf{t}_i] \in SE(3)$, where $\mathbf{R}_i \in SO(3)$ is the rotation and $\mathbf{t}_i \in \mathbb{R}^3$ is the translation.
    \item \textbf{Edges:} Each edge $(i, j) \in \mathcal{E}$ represents a relative pose constraint between nodes $i$ and $j$, where $\mathcal{E}$ is the set of all such connected pairs. Each edge is associated with a relative transformation measurement $\mathbf{Z}_{ij} \in SE(3)$, typically obtained from pairwise pose estimation.
\end{itemize}




The goal of PGO is to find the set of absolute camera poses $\{\mathbf{T}_i\}$ that best explain the observed relative transformations, while ensuring global consistency among all poses throughout the scene. This is formulated as the following optimization problem:

\begin{equation}
\min_{\{\mathbf{T}_i\}} \sum_{(i,j) \in \mathcal{E}} \left\| \mathrm{Log}\left( \mathbf{Z}_{ij}^{-1} \mathbf{T}_i^{-1} \mathbf{T}_j \right) \right\|_{\mathbf{\Omega}_{ij}}^2.
\end{equation}
Here, $\mathrm{Log}(\cdot)$ denotes the logarithmic map from the Lie group $SE(3)$ to its corresponding Lie algebra $\mathfrak{se}(3)$, producing a 6D vector that concatenates rotational and translational residuals. 
The matrix $\mathbf{\Omega}_{ij} \in \mathbb{R}^{6 \times 6}$ is the information matrix, representing the confidence in the measurement $\mathbf{Z}_{ij}$.

This equation jointly optimizes both rotation and translation by aligning the estimated relative transformation $\mathbf{T}_i^{-1} \mathbf{T}_j$ with the measured one $\mathbf{Z}_{ij}$. The residual becomes small when the estimated and measured transformations are close in the $SE(3)$ manifold, thereby promoting global consistency across the entire pose graph.

\vspace{+1mm}
\noindent \textbf{Scene coordinate regression.} Given a query image $I_i$ where $i$ refers to image index, this family of localization methods regresses for a 2D pixel $\mathbf{x}_i \in \mathbb{R}^2$ the corresponding 3D coordinates $\mathbf{X}_i \in \mathbb{R}^3$ in the global scene coordinate frame~\cite{shotton2013scene,brachmann2023accelerated,wang2024glace}. This mapping enables the establishment of 2D--3D correspondences, which can then be used to estimate the camera pose $\mathbf{T}_i = [\mathbf{R}_i | \mathbf{t}_i] \in SE(3)$ via a PnP~\cite{gao2003complete} solver in a RANSAC~\cite{fischler1981random} loop.

Formally, given an image $I_i$
, the scene coordinate regression network $\Phi_\theta$ predicts $\mathbf{X}_i^k = \Phi_\theta(\mathbf{x}_i, I_i).$
The predicted 3D point $\mathbf{X}_i^k$ is assumed to lie in a global coordinate frame shared across all images. Given a set of $K$ such 2D--3D correspondences $\{(\mathbf{x}_i, \mathbf{X}_i)\}_{k=1}^K$ and ground truth camera poses $\mathbf{\hat{T}_{i}}$, the network can be trained by minimizing the reprojection error:
\begin{equation}
 \sum_{k=1}^{K} \left\| \pi(\mathbf{\hat{T}}_i \mathbf{X}_i^k) - \mathbf{x}_i^k \right\|^2,
\end{equation}
where $\pi(\cdot)$ denotes the camera projection function.

\vspace{+1mm}
\noindent \textbf{Learning-based SfM} frameworks typically rely on supervised training, where ground truth poses or depth maps are used to guide the learning of scene geometry and camera motion. ACE-Zero~\cite{brachmann2024acezero} stands out in that it is unsupervised. It recovers camera poses from unordered image collections by iteratively refining pseudo-labels derived from initial pose estimates. The method begins by selecting five random seed images and uses ZoeDepth~\cite{bhat2023zoedepth} to predict depth maps for each. The seed with the highest relocalization rate across 1000 mapping images is chosen to initialize the reconstruction. These depth maps are used to generate initial pseudo-labels, which supervise the training of a scene coordinate regression network. This network is then used to register new images and refine their focal lengths and poses using MLPs. What distinguishes ACE-Zero is its ability to bootstrap accurate scene geometry and camera poses without any external supervision, relying solely on internally generated signals. The system jointly learns scene coordinates and camera poses in a closed loop, where each component improves the other over time. As more images are registered, the scene coordinate regressor is updated and pose estimates are refined, gradually improving the reconstruction without requiring ground truth poses, depth maps, or known 3D geometry. The reconstruction process stops either when 99\% of the images have been successfully registered, or when fewer than 1\% of new images are registered compared to the previous iteration.

ACE-Zero refines camera poses using a learned optimizer that operates locally. Specifically, it updates each pose based only on image-specific information, without enforcing consistency across the entire scene. As a result, the method lacks global constraints and is prone to accumulating drift, leading to inconsistent camera trajectories. To address this limitation, we propose integrating global optimization techniques, such as PGO, which enforce consistency across all estimated poses and improve the overall accuracy and coherence of the reconstructed scene.




\subsection{Pose graph construction} \label{sec:graph_construction}

To enable global pose refinement within the ACE-Zero framework, we construct a pose graph using the predicted scene coordinates and estimated camera poses. The process begins with identifying reliable relative pose constraints between image pairs based on scene coordinate predictions.


At iteration $t$, ACE-Zero predicts dense scene coordinates $\mathbf{X}_i^t \in \mathbb{R}^3$ for each pixel $\mathbf{x}_i^t \in \mathbb{R}^2$ in every image $I_i \in \mathbf{I}$, where $i$ refers to the image index. To establish correspondences between two images $I_i$ and $I_j$, we identify pixel pairs whose predicted scene coordinates are geometrically consistent. Specifically, for each 3D point $\mathbf{X}_i^t$ predicted in $I_i$, we project it into image $I_j$ by first transforming it into the coordinate frame of camera $j$ using the inverse of its absolute pose $\mathbf{T}_j^{-1}$. The projected pixel location is then computed as $\hat{\mathbf{x}}_j^t = \pi(\mathbf{T}_{j}^{-1} \mathbf{X}_i^t)$, where $\pi(\cdot)$ denotes the camera projection function.
At this location in $I_j$, we retrieve the predicted scene coordinate $\mathbf{X}_j^{t}$ and compute the 3D consistency error $e_{\text{match}}^t = \| \mathbf{X}_j^{t} - \mathbf{X}_i^t \|$. If this error is below a predefined threshold $\tau$, the pixel pair $(\mathbf{x}_i^t, \hat{\mathbf{x}}_j^t)$ is considered a spatially consistent match. Optionally, a symmetric check can be performed by projecting $\mathbf{X}_j^{t}$ back into $I_i$ and verifying consistency with $\mathbf{X}_i^t$. These geometrically verified correspondences are then used to estimate the relative transformation between the two camera poses. 



To compute the relative pose $\mathbf{Z}_{ij} = [\mathbf{R}_{ij} | \mathbf{t}_{ij}] \in SE(3)$ between images $I_i$ and $I_j$, we use RANSAC with PnP to find the transformation that best aligns the matched 3D points by minimizing the reprojection error:

\begin{equation}
\mathbf{X}_j^t \approx \mathbf{R}_{ij} \mathbf{X}_i^t + \mathbf{t}_{ij}.
\end{equation}
Once a sufficient number of reliable relative pose estimates are obtained, we construct the pose graph. Each node in the graph corresponds to a camera pose $\mathbf{T}_i$, and each edge $(i, j)$ represents a relative transformation $\mathbf{Z}_{ij}$ between two images. The resulting graph captures the spatial relationships between camera views and serves as the foundation for global pose optimization.

While pose graph construction enables global refinement, it introduces several challenges. The accuracy of the relative pose estimates is sensitive to the quality of the scene coordinate predictions and the geometric consistency threshold $\tau$. A low threshold may reject valid correspondences, while a high threshold risks incorporating noisy matches. Additionally, although RANSAC with PnP is used to robustly estimate relative transformations from matched points, its success depends heavily on the presence of a sufficient number of inliers. In cases where the correspondences are sparse or noisy, it may fail to produce a reliable estimate, even when using only geometrically verified matches. These factors can lead to erroneous or missing edges in the pose graph, which in turn affect the quality of global optimization. To address these limitations, we incorporate uncertainty-aware mechanisms in the subsequent optimization stage to improve robustness and reliability.


\subsection{Uncertainty-aware pose graph optimization}



Incorporating uncertainty into pose graph optimization has been shown to improve robustness and accuracy, particularly in scenarios where input data may be noisy or ambiguous. By weighting constraints according to their confidence, uncertainty-aware methods can down-weight unreliable measurements and prevent them from disproportionately influencing the global solution. In traditional SfM or SLAM pipelines, uncertainty estimates are often derived from sensor noise models, reprojection errors, or covariance propagation during bundle adjustment. However, ACE-Zero does not provide explicit uncertainty estimates for its predicted scene coordinates or camera poses. This presents a challenge for integrating uncertainty into the optimization process. To address this, we propose two unsupervised strategies for estimating confidence in the predicted scene coordinates based on geometric priors. These confidence scores are then used to weight the relative pose constraints in the pose graph, allowing the optimizer to account for the varying reliability of different correspondences.







\begin{figure*}[ht]
\centering
\includegraphics[width=1\linewidth]{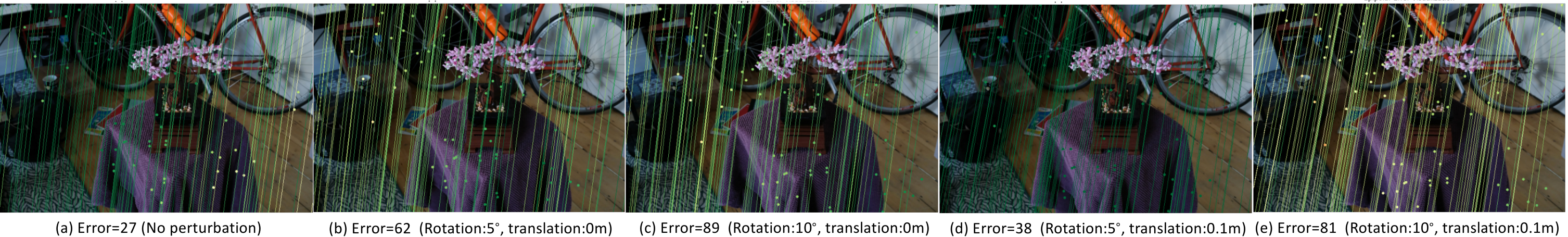}
\caption{Qualitative example of the epipolar-based uncertainty metric. We overlay the reprojected points and their corresponding epipolar lines which are color-coded by error (green=low, red=high), indicating geometric alignment quality under perturbed pose. In (a), errors remain low with no perturbation applied. Introducing pose perturbations in (b) and (c) naturally increases the error. Interestingly, adding translation perturbations in (d) and (e) can sometimes reduce the error, indicating that epipolar error is sensitive to interactions between pose components and does not always correlate directly with pose accuracy. Consequently, we interpret it as a heuristic confidence indicator rather than a strict geometric validator. 
}
\vspace{-0.2cm}
\label{fig:epipolar}
\end{figure*} 

\vspace{+1mm}
\noindent \textbf{Epipolar-based uncertainty metric.}
We use epipolar geometry as a geometric prior to estimate the uncertainty of predicted scene coordinates. Epipolar geometry leverages the fundamental constraint that a 3D point observed in one image must project onto the corresponding epipolar line in another image, given the relative camera poses.

We first project $\mathbf{X}_i^t$ into image $I_j$ to obtain the pixel location $\hat{\mathbf{x}}_j^t = \pi(\mathbf{T}_j^{-1} \mathbf{X}_i^t)$. Using the relative pose, we compute the essential matrix:
\begin{equation}
\mathbf{E}_{ij} = [\mathbf{t}_{ij}]_\times \mathbf{R}_{ij},
\end{equation}

\noindent where $[\mathbf{t}_{ij}]_\times$ is the skew-symmetric matrix of the relative translation vector. Using the camera intrinsic matrix $\mathbf{K}_{j}$, the epipolar line $\mathbf{l}_j^t$ in image $I_j$ corresponding to pixel $\mathbf{x}_i^t$ in image $I_i$ is given by:

\begin{equation}
\mathbf{l}_j^t = \mathbf{K}_{j}^{-\top} \mathbf{E}_{ij} \mathbf{K}_{j}^{-1} \mathbf{x}_i^t.
\end{equation}

We then compute the epipolar uncertainty as the distance between the projected point $\hat{\mathbf{x}}_j^k$ and the epipolar line $\mathbf{l}_j^k$:

\begin{equation}
e_{\text{epi}}^t = \frac{|\mathbf{l}_j^{t\top} \hat{\mathbf{x}}_j^t|}{\sqrt{l_{j,1}^2 + l_{j,2}^2}},
\end{equation}

\noindent where $l_{j,1}, l_{j,2}$ are the first two components of the epipolar line $\mathbf{l}_j^k = [a, b, c]^\top$. The numerator computes the algebraic distance from the point to the line, and the denominator normalizes the line direction to yield the Euclidean (perpendicular) distance from the point to the epipolar line.

This error measures how well the predicted 3D point aligns with the epipolar constraint across views. A lower error indicates higher geometric consistency and thus greater confidence in the scene coordinate prediction. Unlike depth consistency, which compares predicted 3D coordinates directly, epipolar geometry provides a view-invariant constraint that is particularly useful when depth predictions are noisy or ambiguous.


\vspace{+1mm}
\noindent \textbf{Optical flow-based uncertainty metric.}
While epipolar geometry provides global geometric prior for assessing scene coordinate confidence, its effectiveness depends on the accuracy of both the predicted 3D coordinates and the relative camera poses. Errors in scene coordinates can lead to unstable reprojections, and pose inaccuracies distort the essential matrix, resulting in misleading epipolar lines--particularly during early training stage. 


As shown in Fig.~\ref{fig:epipolar}, the epipolar-based uncertainty metric is often weakly sensitive to small pose perturbations. Even when camera poses deviate from ground truth, the resulting epipolar residuals may remain low. This suggests that epipolar metric alone may not be sufficiently discriminative for uncertainty estimation. While it can indicate inconsistency, it may also underrepresent uncertainty in ambiguous or noisy conditions. 

\begin{figure}[t]
\centering
\includegraphics[width=0.85\linewidth]{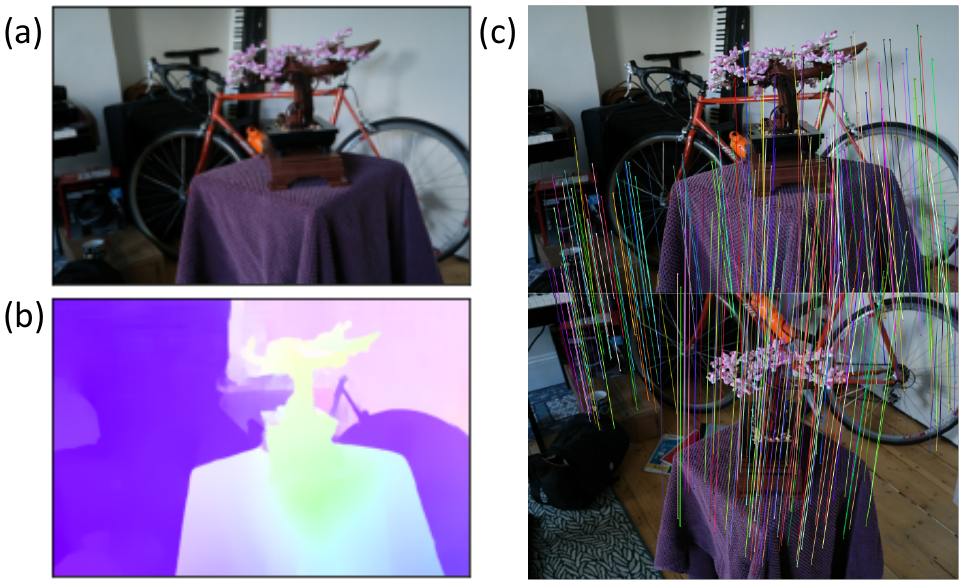}
\caption{Qualitative example of optical flow predictions. (a) One image from the input pair. (b) Predicted flow visualization where similar colours indicate similar motion directions. (c) UV correspondences overlaid on the image pair, demonstrating that optical flow provides robust pixel-level matches between views.}
\vspace{-0.2cm}
\label{fig:optical}
\end{figure}

To address these limitations, we introduce a complementary metric based on optical flow, which provides local and image-driven evidence that is more robust to pose inaccuracies and complements the global nature of epipolar geometry. As shown in Fig.~\ref{fig:optical}, dense optical flow provides pixel-level correspondences across views that are robust to pose errors. Unlike epipolar geometry, which depends on global pose accuracy, optical flow captures local motion patterns directly from image content.
Specifically, we obtain the dense optical flow field $\mathbf{F}_{i \rightarrow j}$ from RAFT~\cite{teed2020raft}, which estimates the pixel displacement between $I_i$ and $I_j$. The flow-predicted correspondence of $\mathbf{x}_i^t$ in $I_j$ is:
\begin{equation}
\tilde{\mathbf{x}}_j^t = \mathbf{x}_{i}^t + \mathbf{F}_{i \rightarrow j}(\mathbf{x}_i^t).
\end{equation}
We then define the optical flow consistency error as the Euclidean distance between the projected point $\hat{\mathbf{x}}_j^t$ and the flow-predicted correspondence $\tilde{\mathbf{x}}_j^t$:
\begin{equation}
e_{\text{flow}}^t = \left\| \hat{\mathbf{x}}_j^t - \tilde{\mathbf{x}}_j^t \right\|_2.
\end{equation}
This error measures how well the predicted 3D point aligns with the pixel-level motion observed in the image pair. A lower error indicates that the reprojection is consistent with the observed optical flow, suggesting higher confidence in the scene coordinate prediction.

By incorporating both epipolar and optical flow uncertainty metrics, we obtain a more robust confidence estimation framework. Epipolar constraints provide global geometric validation, while optical flow offers local, image-driven evidence. Together, they improve the reliability of uncertainty-aware pose graph optimization in ACE-Zero.

\vspace{+1mm}
\noindent \textbf{Combined confidence score.}
To model the uncertainty of each relative pose constraint in the pose graph, we compute a confidence score for every matched pixel pair between two images. For a pixel $\mathbf{x}_i^t$ in image $I_i$, its confidence score $\sigma_{ij}^t$ is defined as a weighted sum of two geometric error metrics:
\begin{equation}
\sigma_{ij}^t = \alpha_1 e_{\text{epi}}^t + \alpha_2 e_{\text{flow}}^t,
\end{equation}
where $\alpha_1$, $\alpha_2$ are hyperparameters. These scores reflect the geometric reliability of each match and are then aggregated over matched points to compute edge weights in the pose graph optimization.

For each image pair $(i, j)$, we compute the edge-level uncertainty $\sigma_{ij}$ as the mean of all valid per-match confidence scores.
This aggregated uncertainty is then used to define the standard deviation of the noise model for the relative pose constraint in PGO. Specifically, we construct the information matrix $\Omega_{ij}$ as:
\begin{equation}
\Omega_{ij} = \frac{1}{\sigma_{ij}^2 + \epsilon} \cdot I,
\end{equation}
where $\epsilon$ is a small constant added for numerical stability, and $I$ is the identity matrix. This diagonal information matrix allows the optimizer to treat constraints with lower confidence as noisier, thereby reducing their influence during global refinement.

\subsection{Global optimization}

Once the pose graph has been constructed and the edge weights have been adjusted based on uncertainty estimates, we perform global optimization to refine the camera poses. Specifically, we apply pose graph optimization at every iteration of the ACE-Zero framework to continuously improve the global consistency of the estimated trajectory. This iterative refinement ensures that local updates to scene coordinates and camera poses are propagated throughout the entire graph, mitigating drift and reinforcing spatial coherence.

By integrating pose graph optimization into each iteration of ACE-Zero, we enable consistent global refinement that complements local updates, resulting in improved accuracy and stability of the reconstructed camera trajectory.
\section{Experiments}

\vspace{+1mm}
\noindent \textbf{Implementation details.} Our pipeline is based on the public implementation of ACE-Zero~\cite{brachmann2024acezero} and all hyperparameters are kept in their default settings. All experiments are initialized using ZoeDepth~\cite{bhat2023zoedepth} and executed on a single NVIDIA 3090 GPU. We initialize the geometric consistency threshold at $\tau=0.2\text{m}$ and progressively reduce it to $0.05\text{m}$ over training iterations to enforce stricter correspondence checks as predictions become more accurate.
We set $\alpha_1 = 0.4$ and $\alpha_2 = 0.6$, and scale the information matrix using the median confidence value for numerical stability during optimization.
During pose graph optimization, a tight prior is applied to the first pose using a noise model with a standard deviation of $10^{-6}$ for all components. Between factors are modeled with standard deviation of $0.1$ for rotational and translational components. To improve robustness against outliers, we employ a Huber loss function with a scale parameter of $1.0$, which reduces the influence of large residuals while preserving sensitivity to small errors. For nodes that remain unconnected after edge filtering, we add loose priors using a standard deviation of $10$ for all components to prevent optimization failure without introducing strong constraints. The final optimization is performed using Levenberg--Marquardt optimizer.

\vspace{+1mm}
\noindent \textbf{Baselines.} We treat COLMAP-generated poses as pseudo ground truth. Besides default settings, we also report results using the recommended parameter settings for large image collections, which we denote as COLMAP (fast). We compare against our most relevant method, ACE-Zero (ACE0)~\cite{brachmann2024acezero} as well as two neural SLAM approaches, \ie DRIOD-SLAM~\cite{teed2021droid} and VGGT-SLAM~\cite{maggio2025vggt}.

\vspace{+1mm}
\noindent \textbf{Datasets.} We follow the evaluation protocol from \cite{brachmann2024acezero} and experiment on three benchmarks: 7-Scenes~\cite{shotton2013scene} for relocalization, Mip-NeRF 360~\cite{barron2022mip} for view synthesis, and Tanks and Temples~\cite{knapitsch2017tanks} for 3D reconstruction.

\vspace{+1mm}
\noindent \textbf{Evaluation metrics.} As obtaining ground truth camera poses in real-world settings is challenging, pseudo ground truth camera poses are estimated rather than measured. Consequently, some methods may outperform COLMAP in certain scenes which aligns with prior findings that pose error computed against COLMAP can be misleading. Following established methods~\cite{brachmann2024acezero,kerbl20233d}, we assess pose quality in a self-supervised manner using novel view synthesis~\cite{waechter2017virtual}. Specifically, each method estimates poses for all images in a scene. We then split the images into training and test sets, train a Nerfacto~\cite{tancik2023nerfstudio} model on the training set, and synthesize views for the test poses. The synthesized images are compared to the test images using peak signal-to-noise ratio (PSNR). Pose quality influences PSNR in two ways: accurate training poses enable the NeRF model to learn a consistent scene representation, while accurate test poses ensure alignment between synthesized and original views.

\subsection{Experimental results}

\vspace{+1mm}
\noindent \textbf{7-Scenes (Tab.~\ref{tab:sevenscenes}).} This dataset consists seven indoor scenes. It provides pseudo ground-truth camera poses estimated by KinectFusion~\cite{izadi2011kinectfusion}, a depth-based SLAM system. Although KinectFusion operates in real time, its poses yield lower PSNR; thus, we omit these results from our tables and refer readers to \cite{brachmann2024acezero} for reported numbers. Here, we use COLMAP poses as the reference pseudo ground truth. We show that our approach matches or surpass the performance of both COLMAP and ACE-Zero while requiring significantly less computation time. 

\begin{table}[t]
\bgroup
    \centering%
    \large
    \adjustbox{valign=t,width=\linewidth}{%
        \begin{tabular}{cc||cc||cccccccc}
             & & & & & & & & & & &   \\ 
            \toprule
            \multicolumn{2}{c||}{~} &%
            \multicolumn{2}{c||}{~Pseudo Ground Truth~} &%
            \multicolumn{8}{c}{~All Frames~} 
             \\
            \hhline{~~----------}
             ~ &%
            \hspace{-6pt}\parbox[t]{0mm}{\multirow{2}{*}{\rotatebox[origin=c]{90}{Frames\hspace{-12pt}}}}
            &%
            ~COLMAP~ &%
            ~COLMAP~ &%
            \multicolumn{2}{c}{~ACE0~} &%
            \multicolumn{2}{c}{~Ours~} &%
            \multicolumn{2}{c}{~Ours~} &%
            \multicolumn{2}{c}{~Ours~} 
             \\
             &%
             ~ &%
            (default) &%
            (fast) &%
            \multicolumn{2}{c}{\cite{brachmann2024acezero}} &%
            \multicolumn{2}{c}{(vanilla pgo)} &%
            \multicolumn{2}{c}{(w/o uncertainty)} &%
            \multicolumn{2}{c}{(full)} 
             \\
            \midrule 
             Chess & 6k & {\cellcolor{TableDarkGreen}23.6} & 23.5 & \multicolumn{2}{c}{\begin{tabular}{cc} & ~{23.3 (23.1)}~ \end{tabular}} &  \multicolumn{2}{c}{\begin{tabular}{cc} & ~{11.2}~ \end{tabular}} &  \multicolumn{2}{c}{\begin{tabular}{cc} & ~{19.2}~  \end{tabular}} &  \multicolumn{2}{c}{\begin{tabular}{cc} & ~{23.5}~ \end{tabular}}  \\
             & & & & & & & & & & &  \\[-10pt]
 Fire & 4k & {\cellcolor{TableDarkGreen}22.6} & {\cellcolor{TableDarkGreen}22.6}  & \multicolumn{2}{c}{\begin{tabular}{cc} & ~{22.3 (22.0)}~  \end{tabular}} &  \multicolumn{2}{c}{\begin{tabular}{cc} & ~{13.3}~ \end{tabular}} &  \multicolumn{2}{c}{\begin{tabular}{cc} & ~{19.0}~ \end{tabular}} &  \multicolumn{2}{c}{\begin{tabular}{cc} & ~{\cellcolor{TableDarkGreen}22.6}~ \end{tabular}}  \\
             & & & & & & & & & & & \\[-10pt]
 Heads & 2k & 18.8 & {18.9}  & \multicolumn{2}{c}{\begin{tabular}{cc} & ~{18.8 (18.6)}~ \end{tabular}} &  \multicolumn{2}{c}{\begin{tabular}{cc} & ~{11.9}~ \end{tabular}} &  \multicolumn{2}{c}{\begin{tabular}{cc} & ~{17.7}~ \end{tabular}} &  \multicolumn{2}{c}{\begin{tabular}{cc} & ~{\cellcolor{TableDarkGreen}19.2}~ \end{tabular}}  \\
             & & & & & & & & & & &  \\[-10pt]
 Office & 10k & 21.4 & {\cellcolor{TableDarkGreen}21.6}  & \multicolumn{2}{c}{\begin{tabular}{cc} & ~{21.1 (21.0)}~ \end{tabular}} &  \multicolumn{2}{c}{\begin{tabular}{cc} & {12.4} \end{tabular}} &  \multicolumn{2}{c}{\begin{tabular}{cc} & ~{18.5}~ \end{tabular}} &  \multicolumn{2}{c}{\begin{tabular}{cc} & ~{\cellcolor{TableDarkGreen}21.6}~ \end{tabular}}  \\
             & & & & & & & & & & &  \\[-10pt]
 Pumpkin & 6k & {24.1} & 23.8 & \multicolumn{2}{c}{\begin{tabular}{cc} & ~{24.1 (24.0)}~ \end{tabular}} &  \multicolumn{2}{c}{\begin{tabular}{cc} & ~{14.5}~ \end{tabular}} &  \multicolumn{2}{c}{\begin{tabular}{cc} & ~{19.3}~ \end{tabular}} &  \multicolumn{2}{c}{\begin{tabular}{cc} & ~{\cellcolor{TableDarkGreen}24.3}~ \end{tabular}}  \\
             & & & & & & & & & & & \\[-10pt]
 RedKitchen & 12k & {\cellcolor{TableDarkGreen}21.4} & {\cellcolor{TableDarkGreen}21.4}  & \multicolumn{2}{c}{\begin{tabular}{cc} & ~{20.8 (20.9)}~ \end{tabular}} &  \multicolumn{2}{c}{\begin{tabular}{cc} & ~{11.9}~ \end{tabular}} &  \multicolumn{2}{c}{\begin{tabular}{cc} & ~{18.5}~ \end{tabular}} &  \multicolumn{2}{c}{\begin{tabular}{cc} & ~{21.2}~ \end{tabular}}  \\
             & & & & & & & & & & &  \\[-10pt]
 Stairs & 3k & 16.7 & {\cellcolor{TableDarkGreen}21.0} & \multicolumn{2}{c}{\begin{tabular}{cc} & ~{17.7 (17.3)}~ \end{tabular}} &  \multicolumn{2}{c}{\begin{tabular}{cc} & ~{12.7}~  \end{tabular}} & \multicolumn{2}{c}{\begin{tabular}{cc} & ~{17.5}~  \end{tabular}} &  \multicolumn{2}{c}{\begin{tabular}{cc} & ~{20.1}~  \end{tabular}}  \\
             & & & & & & & & & & & \\[-10pt]
\hhline{~-----------}
             & & & & & & & & & & &  \\[-10pt]
             Average &      & 21.2 & {\cellcolor{TableDarkGreen}21.8} & \multicolumn{2}{c}{\begin{tabular}{cc} & ~{21.2 (21.0)}~  \end{tabular}} &  \multicolumn{2}{c}{\begin{tabular}{cc} & {12.6}  \end{tabular}} &  \multicolumn{2}{c}{\begin{tabular}{cc} & ~{18.5}~  \end{tabular}} &  \multicolumn{2}{c}{\begin{tabular}{cc} & ~{21.7}~  \end{tabular}}  \\
             & & & & & & & & & & & \\[-10pt]
\hline
             Avg. Time &      & 38h & 13h & \multicolumn{2}{c}{\begin{tabular}{cc} & ~1h~  \end{tabular}} &  \multicolumn{2}{c}{\begin{tabular}{cc} & ~1h~  \end{tabular}} &  \multicolumn{2}{c}{\begin{tabular}{cc} & ~1h~  \end{tabular}} &  \multicolumn{2}{c}{\begin{tabular}{cc} & ~30min~  \end{tabular}} \\
             & & & & & & & & & & & \\[-10pt]
\bottomrule
      \end{tabular}
    }
    \caption{
        Quantitative comparisons on 7-Scenes dataset.
        We report the pose accuracy via view synthesis as PSNR in dB and the reconstruction time. \TCBTableDarkGreen{Dark green} indicate the best results. Values shown in parentheses correspond to our self-reproduction results. 
    }%
    \label{tab:sevenscenes}%
\vspace{-0.6cm}
\egroup
\end{table}

\vspace{+1mm}
\noindent \textbf{Mip-NeRF 360 (Tab.~\ref{tab:mip}).}  This dataset consists of seven small-scale scenes which covers both indoor and outdoor. Similarly, we use COLMAP generated poses as the reference pseudo ground truth. VGGT-SLAM~\cite{maggio2025vggt} is a recent dense RGB SLAM system constructed by incrementally and globally aligning submaps generated from VGGT~\cite{wang2025vggt}. We observe that VGGT-SLAM struggles in this dataset due to the large differences between sequential images. In contrast, ACE-Zero successfully reconstructs the scenes and achieves relatively better performance. Nevertheless, our approach demonstrates improvements over ACE-Zero while maintaining competitive results against COLMAP.

\begin{table}[t]
\bgroup
    \centering%
    \large
    \adjustbox{valign=t,width=\linewidth}{%
        \begin{tabular}{ccccccc}
             & & & & & &    \\ 
            \toprule
             ~ &%
            
            ~COLMAP~ &%
            ~VGGT-~ &%
            {~ACE0~} &%
            {~Ours~} &%
            {~Ours~} &%
            {~Ours~} 
             \\
             &%
             ~ 
            (default) &%
            SLAM~\cite{maggio2025vggt} &%
            {\cite{brachmann2024acezero}} &%
            {(vanilla pgo)} &%
            {(w/o uncertainty)} &%
            {(full)} 
             \\
            \midrule 
             Bicycle  & {21.5} & {11.8} & {18.7 (18.3)} &   ~{12.5}~  &   ~{19.0}~  &   {\cellcolor{TableDarkGreen}21.9}  \\
             & & & & & &   \\[-10pt]
 Bonsai & {\cellcolor{TableDarkGreen}27.6} & {16.7}  & {25.8 (25.5)} &   ~{13.9}~  &   ~{22.1}~  &  ~{26.6}~  \\
              & & & & & &  \\[-10pt]
 Counter  & {\cellcolor{TableDarkGreen}25.5} & {12.8}  &  ~{24.5 (24.8)}&  ~{14.5}~  &  ~{21.2}~  &  ~{25.2}~   \\
              & & & & & &   \\[-10pt]
 Garden & {26.3} & {17.9}  &  ~{25.0 (24.5)}~  &   {14.7}  &   ~{22.5}~  &   ~{\cellcolor{TableDarkGreen}26.6}~   \\
              & & & & & &   \\[-10pt]
 Kitchen & {\cellcolor{TableDarkGreen}27.4} & {14.3} &  ~{26.1 (26.3)}~  &   ~{14.3}~  &   ~{23.1}~ &   ~{26.8}~   \\
              & & & & & &  \\[-10pt]
 Room  & {\cellcolor{TableDarkGreen}28.0} & {12.6}  &  ~{19.8 (19.5)}~  &   ~{14.3}~  &   ~{19.3}~  &  ~{21.8}~   \\
              & & & & & &   \\[-10pt]
 Stump & 16.8 & {13.7} &  ~{20.5 (16.1)}~  &   ~{12.7}~   &  ~{20.1}~   &   ~{\cellcolor{TableDarkGreen}21.4}~   \\
              & & & & & &  \\[-10pt]
\hline
             &  & & & & &   \\[-10pt]
             Average &      {\cellcolor{TableDarkGreen}24.7} & {14.3} &  ~{22.9 (22.14)}~   &   {13.8}   &   ~{21.0}~   &   ~{24.3}~  \ \\
              & & & & & &  \\[-10pt]
\bottomrule
      \end{tabular}
    }
    \caption{
        Quantitative comparisons on Mip-NeRF 360 dataset.
        We report the pose accuracy via view synthesis as PSNR in dB and the reconstruction time. \TCBTableDarkGreen{Dark green} indicate the best results. Values shown in parentheses correspond to our self-reproduction results. 
    }%
    \label{tab:mip}%
\vspace{-0.2cm}
\egroup
\end{table}

\begin{figure*}[t]
\centering
\includegraphics[width=0.89\linewidth]{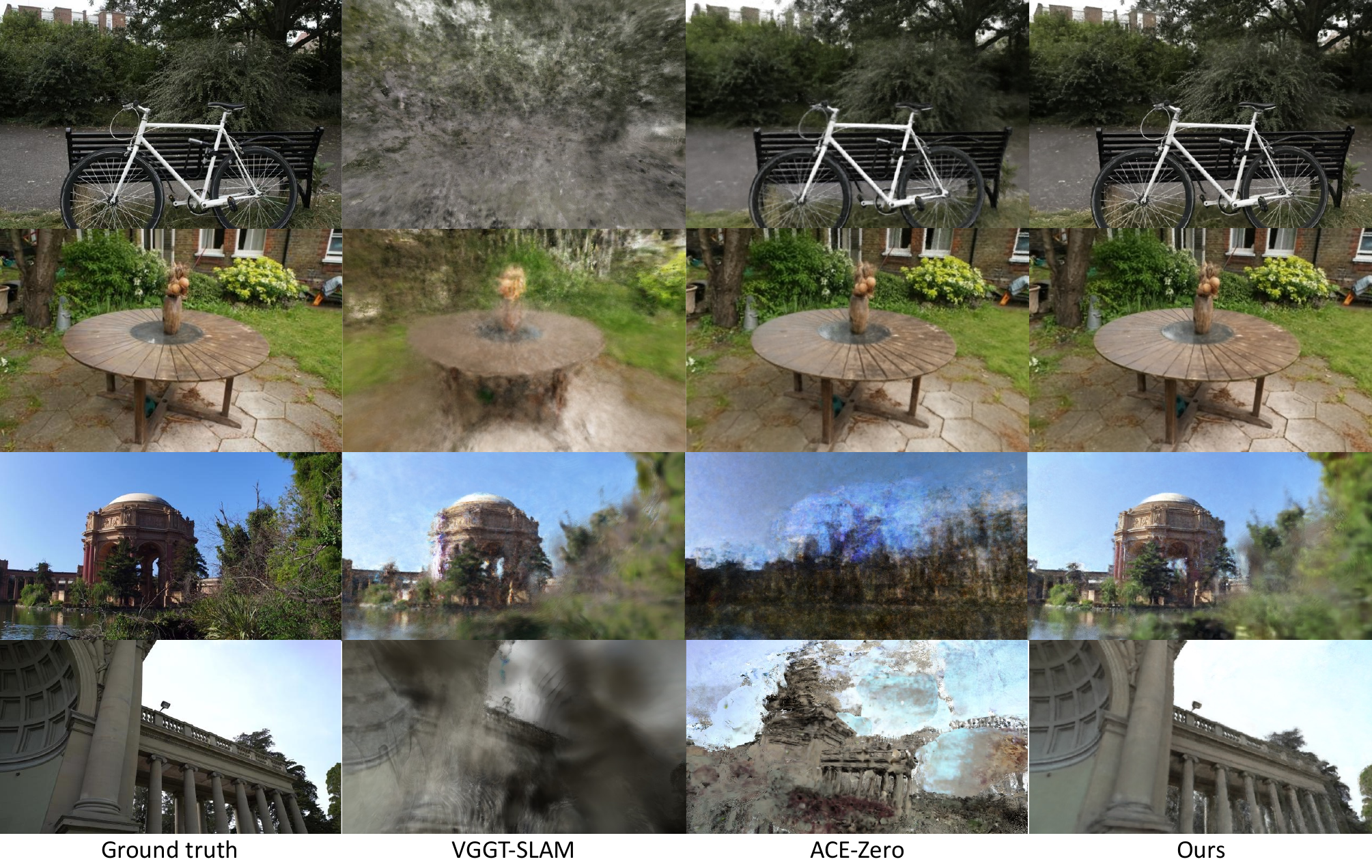}
\caption{Qualitative comparison on Mip-NeRF 360 (top two rows) and Tanks and Temples datasets (bottom two rows) with VGGT-SLAM~\cite{maggio2025vggt} and ACE-Zero~\cite{brachmann2024acezero}. While ACE-Zero performs well on small-scale scenes such as Mip-NeRF 360, it struggles with large, complex environments like Tanks and Temples, where our method maintains high rendering quality across diverse scenes.
}
\vspace{-0.2cm}
\label{fig:compairson}
\end{figure*}

\vspace{+1mm}
\noindent \textbf{Tanks and Temples (Tab.~\ref{tab:t2}).} This dataset comprises 21 diverse indoor and outdoor scenes, covering a substantially larger spatial extent compared to the previous two datasets. We present results for both the short version of the dataset (150–500 images) and the long version (4k–22k images). Across both versions, our proposed method generally outperforms Ace-Zero and achieves performance comparable to COLMAP. Notably, on the Advanced split, VGGT-SLAM delivers similar accuracy to ACE-Zero (13.6 vs. 14.5 PSNR on average) while requiring significantly less reconstruction time. For the long version, we attempt to downsample sequence to around 4k frames for VGGT-SLAM in order fit GPU memory. However, we observe fracture scenes for all sequences, so their performance is not reported. We provide qualitative comparisons in Fig.~\ref{fig:compairson}.

\begin{table}[t]%
    \centering%
    \setlength{\tabcolsep}{1pt}%
    \footnotesize%
    \adjustbox{valign=t,width=1\linewidth}{%
} 
         \\
        \bottomrule
    \end{tabular}
    }
    \caption{Quantitative comparisons on Tanks and Temples dataset. We report the pose accuracy via view synthesis as PSNR in dB and the reconstruction time. \TCBTableDarkGreen{Dark green} indicate the best results, \textbf{excluding} COLMAP.
    }%
    \label{tab:t2}%
\end{table}

\begin{figure}
\centering
\includegraphics[width=1\linewidth]{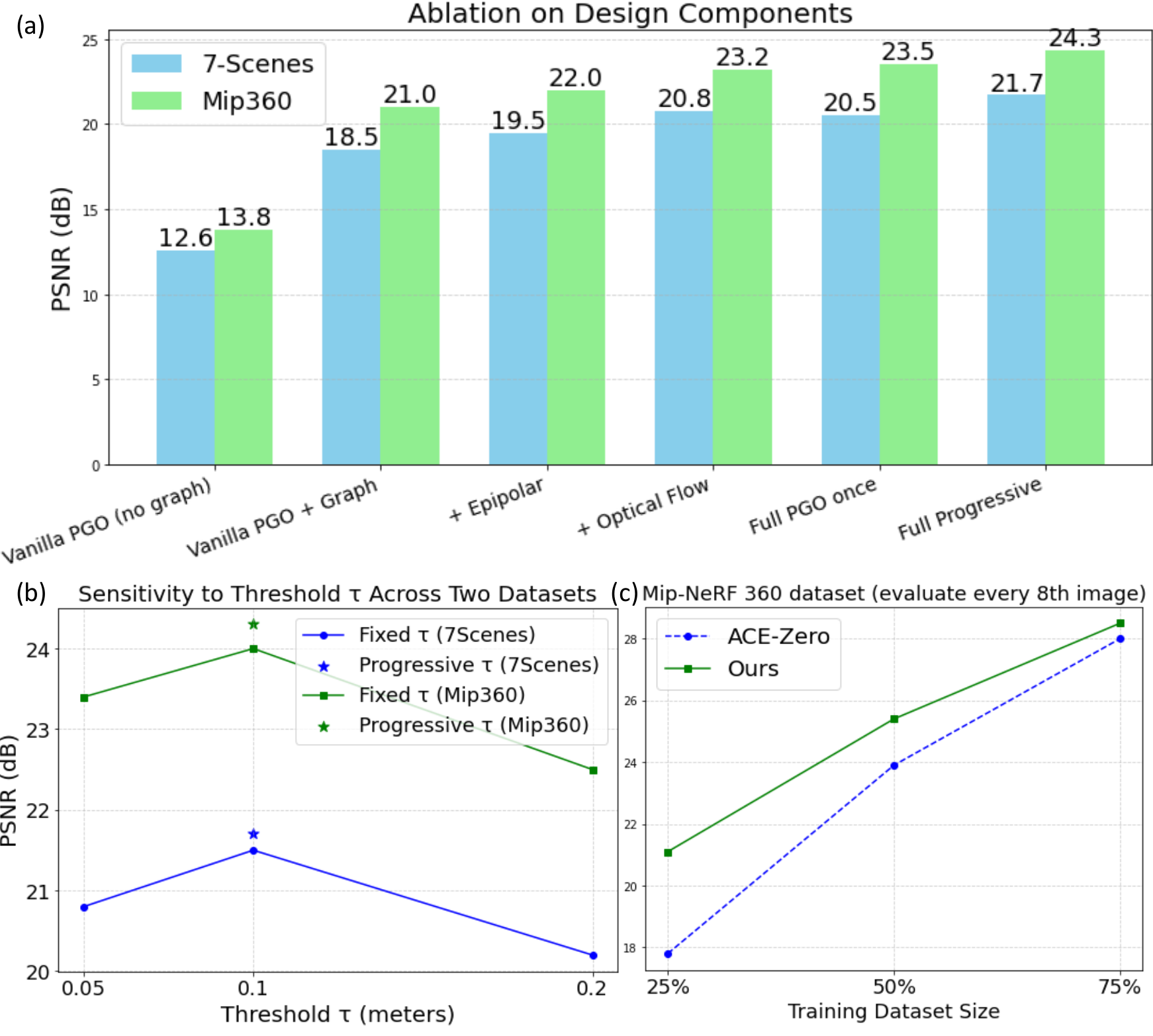}
\caption{Ablations on (a) design components, and sensitivity to (b) hyperparameter and (c) sparse training data.
}
\vspace{-0.25cm}
\label{fig:ablation}
\end{figure}

\subsection{Ablations}
\vspace{+1mm}
\noindent \textbf{Pose graph optimization (Fig.~\ref{fig:ablation}(a)).} We ablate key components of our pipeline on 7-Scenes and Mip-NeRF 360 datasets. Vanilla PGO without graph construction performs poorly due to unreliable edges (12.6dB, 13.8dB). Adding graph construction significantly improves accuracy ($>$5dB), and incorporating epipolar-based uncertainty yields moderate gains. Optical flow further boosts robustness by providing local consistency. Applying PGO only once at the end underperforms compared to progressive integration (20.5dB vs. 21.7dB; 23.5dB vs. 24.3dB), highlighting the importance of iterative global refinement.

\vspace{+1mm}
\noindent \textbf{Pose graph construction (Fig.~\ref{fig:ablation}(b)).} 
We evaluate the impact of fixing the geometric consistency threshold $\tau$ at 0.05m, 0.1m and 0.2m. We show that $\tau=0.1\text{m}$ achieves the best performance among fixed settings, balancing accuracy and graph connectivity, while $\tau=0.2\text{m}$ admits noisy matches that degrade global consistency. We also tested enforcing bidirectional consistency when constructing pose graph edges. Symmetric checks slightly improve PSNR in dense indoor scenes with rich texture, as they yield more reliable edges. However, the gains are modest ($<5\%$) overall and performance can degrade in sparser settings because stricter filtering reduces graph connectivity, leading to fewer edges and weaker global consistency. Given the limited benefit and added complexity, we adopt one-way checks in our default pipeline.


\vspace{+1mm}
\noindent \textbf{Impact on less input images (Fig.~\ref{fig:ablation}(c)).} Our method demonstrates better robustness to limited training data.

\vspace{+1mm}
\noindent \textbf{Runtime breakdown.} Our added PGO step takes about 30s per iteration, compared to 150s for vanilla ACE-Zero. Despite the extra computation, PGO reduces the total number of iterations by roughly half. We provide detailed breakdown in the supplementary material.

\section{Conclusion}
We presented a hybrid framework that integrates pose graph optimization into ACE-Zero for learning scene coordinate reconstruction from unposed images. By progressively enforcing global consistency and incorporating uncertainty-aware weighting, our approach significantly improves pose accuracy and robustness across diverse datasets. This work demonstrates that combining neural inference with geometric reasoning yields scalable and reliable reconstructions. Despite strong results, our method has several limitations. First, uncertainty estimation relies on heuristic metrics, which may not generalize well to highly dynamic or textureless scenes. Second, the hyperparameters requires manual tunning. Third, while our approach improves global consistency, it does not jointly optimize scene structure, leaving room for future work on full bundle adjustment.
\clearpage
{
    \small
    \bibliographystyle{ieeenat_fullname}
    \bibliography{main}
}

\clearpage
\section{Appendix}
\appendix
In this supplemental document, we provide:
\begin{itemize}
    \item extended implementation details and discussion (Sec~\ref{sup:implementation});
    \item additional analysis (Sec~\ref{sup:analysis});
    \item additional qualitative examples (Sec~\ref{sup:qualitative}).
\end{itemize}

\section{Extended implementation details and discussion} \label{sup:implementation}

\paragraph{Overall algorithm.} We present the complete pseudocode algorithm in Algorithm~\ref{alg:hybrid}. Our key contribution is to suppress incorrectly refined camera poses, enable a more accurate and globally consistent set of poses. This improved alignment allows the scene coordinate regression network to be trained on a more reliable pose set in subsequent iterations. Consequently, the overall reconstruction process converges faster by reaching accurate camera poses earlier, supported by pose graph optimization.
\begin{algorithm}[t]
\caption{Overall algorithm}
\label{alg:hybrid}
\KwIn{Unordered image set $\mathcal{I} = \{I_1, I_2, \dots, I_n\}$}
\KwOut{Refined camera poses $\{T_i\}$ and trained scene coordinate regression network $\Phi_\theta$}

\textbf{Initialization:}\\
\quad Select seed images and predict depth maps using ZoeDepth.\\
\quad Generate initial pseudo-labels for scene coordinates.\\

\textbf{Iterative Reconstruction Loop:}\\
\For{each iteration $t$}{
    \textbf{(1) Model training}\\
    \quad Train scene coordinate regression network $\Phi_\theta$ on current set of images with pseudo-poses.\\

    \textbf{(2) Pose estimation with previously trained model}\\
    \quad Use trained scene coordinate regression network $\Phi_\theta$ to estimate poses for all images.\\
    \quad Update pseudo-poses for unregistered images.\\

    \textbf{(3) Scene coordinate prediction}\\
    \quad Predict dense scene coordinates $X_i = \Phi_\theta(I_i)$ for each image.\\

    \textbf{(4) Pose Graph Construction}\\
    \For{each image pair $(I_i, I_j)$}{
        Find geometrically consistent matches: project $X_i$ into $I_j$, compute $e_{\text{match}} = \|X_j - X_i\|$, accept if $e_{\text{match}} < \tau$.\\
        Estimate relative pose $Z_{ij}$ using PnP + RANSAC.\\
        Add edge $(i,j)$ with $Z_{ij}$ to pose graph.\\
    }

    \textbf{(5) Uncertainty estimation}\\
    \quad Compute epipolar error $e_{\text{epi}}$ and optical flow error $e_{\text{flow}}$.\\
    \quad Confidence score $\sigma = \alpha_1 e_{\text{epi}} + \alpha_2 e_{\text{flow}}$.\\
    \quad Aggregate $\sigma$ to edge-level uncertainty $\sigma_{ij}$ and compute $\Omega_{ij} = \text{diag}(1 / (\sigma_{ij}^2 + \epsilon))$.\\

    \textbf{(6) Pose Graph Optimization}\\
    \quad Solve $\min_{\{T_i\}} \sum_{(i,j)\in E} \|\text{Log}(Z_{ij}^{-1} T_i^{-1} T_j)\|^2_{\Omega_{ij}}$ using LM optimizer with Huber loss.\\

    \textbf{(7) Check stopping criteria}\\
    \quad Stop if $99\%$ images registered or $<1\%$ new.\\
}

\end{algorithm}

\paragraph{Impact of graph sparsity or connectivity on optimization quality.}
Pose graph connectivity plays a critical role in the success of global optimization. A sparse graph, where only a few reliable edges exist, limits the propagation of global constraints and can lead to fragmented or weakly consistent pose estimates. Conversely, an overly dense graph may include noisy or incorrect edges, which can degrade optimization quality by introducing conflicting constraints. We observed that:
\begin{itemize}
    \item \textbf{Sparse Graphs:} When the geometric consistency threshold $\tau$ is set too low, many valid correspondences are rejected, resulting in fewer edges. This reduces the optimizer’s ability to enforce global consistency, especially in large-scale scenes with repetitive structures.
    \item \textbf{Dense Graphs:} A high $\tau$ admits more matches, increasing connectivity but also introducing outliers. These noisy edges can bias the optimization, even when robust loss functions are applied.
    \item \textbf{Balanced Connectivity:} Our experiments (see Fig.~5(b) in the main paper) show that $\tau = 0.1\,\text{m}$ achieves the best trade-off between accuracy and connectivity. This setting provides sufficient edges for global consistency while minimizing the inclusion of unreliable matches.
\end{itemize}

\paragraph{Can the uncertainty-aware metrics provide supervision signal to model training?}
Our current framework uses uncertainty-aware metrics exclusively for weighting constraints during pose graph optimization. These metrics---epipolar error and optical flow consistency---quantify geometric reliability between image pairs without requiring ground truth poses. This raises an interesting question: can these metrics also serve as supervision signals for training the ACE model? We summarize the potential benefits:
\begin{itemize}
    \item \textbf{Self-Supervised Learning:} Epipolar and flow-based consistency could act as additional loss terms, encouraging the ACE model to predict scene coordinates that satisfy multi-view geometric constraints.
    \item \textbf{Improved Robustness:} Incorporating these metrics during training may reduce reliance on pseudo-labels and improve performance in ambiguous or textureless regions.
    \item \textbf{Dynamic Weighting:} Confidence scores could guide curriculum learning, where reliable correspondences are emphasized early in training.
\end{itemize}

One major challenge is non-differentiability: epipolar distance and optical flow errors involve discrete matching and RANSAC steps, making direct backpropagation difficult. Additionally, these metrics depend on estimated poses, which may be inaccurate during early iterations, potentially introducing bias into the training process. Finally, computing optical flow and epipolar constraints for all image pairs during training would significantly increase computational overhead, making the approach less practical for large-scale datasets.

Future work could explore differentiable approximations of epipolar and flow consistency, enabling their integration into end-to-end training. Another promising direction is to incorporate confidence-based weighting into ACE’s reprojection loss, allowing uncertainty-aware metrics to influence optimization directly. Finally, curriculum strategies could be investigated, where uncertainty-aware metrics gradually gain importance as pose estimates become more reliable, balancing computational cost and training stability.


\section{Additional analysis} \label{sup:analysis} 

\begin{figure*}[t]
\centering
\includegraphics[width=1\linewidth]{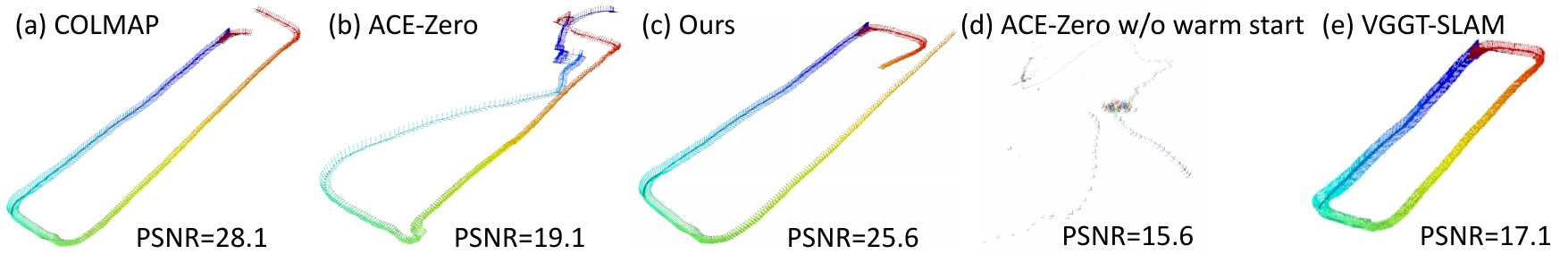}
\caption{Despite being initialized from (a) COLMAP generated camera poses, local refinements in (b) often introduce global inconsistencies. In contrast, (c) our hybrid-framework, which integrates pose graph optimization into ACE-Zero, enforces multi-view consistency and suppress incorrect refinements. Furthermore, we demonstrate that (d) ACE-Zero fails without COLMAP initialization, and (e) the low PSNR indicates that VGGT-SLAM lacks accuracy despite its ability to perform loop closure. 
}
\label{sup:fig:teaser}
\end{figure*}

\begin{figure*}[h]
\centering
\includegraphics[width=1\linewidth]{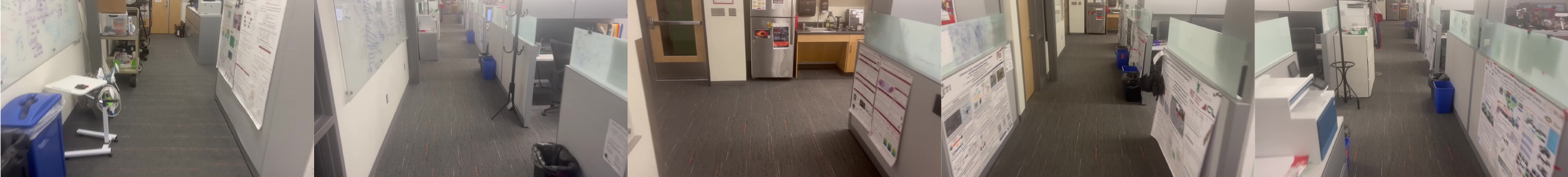}
\caption{We provide several sample image frames from the \textit{office loop} video that were used for the teaser example.
}
\label{sup:fig:office_loop_frames}
\end{figure*}

\begin{figure*}[h]
\centering
\includegraphics[width=1\linewidth]{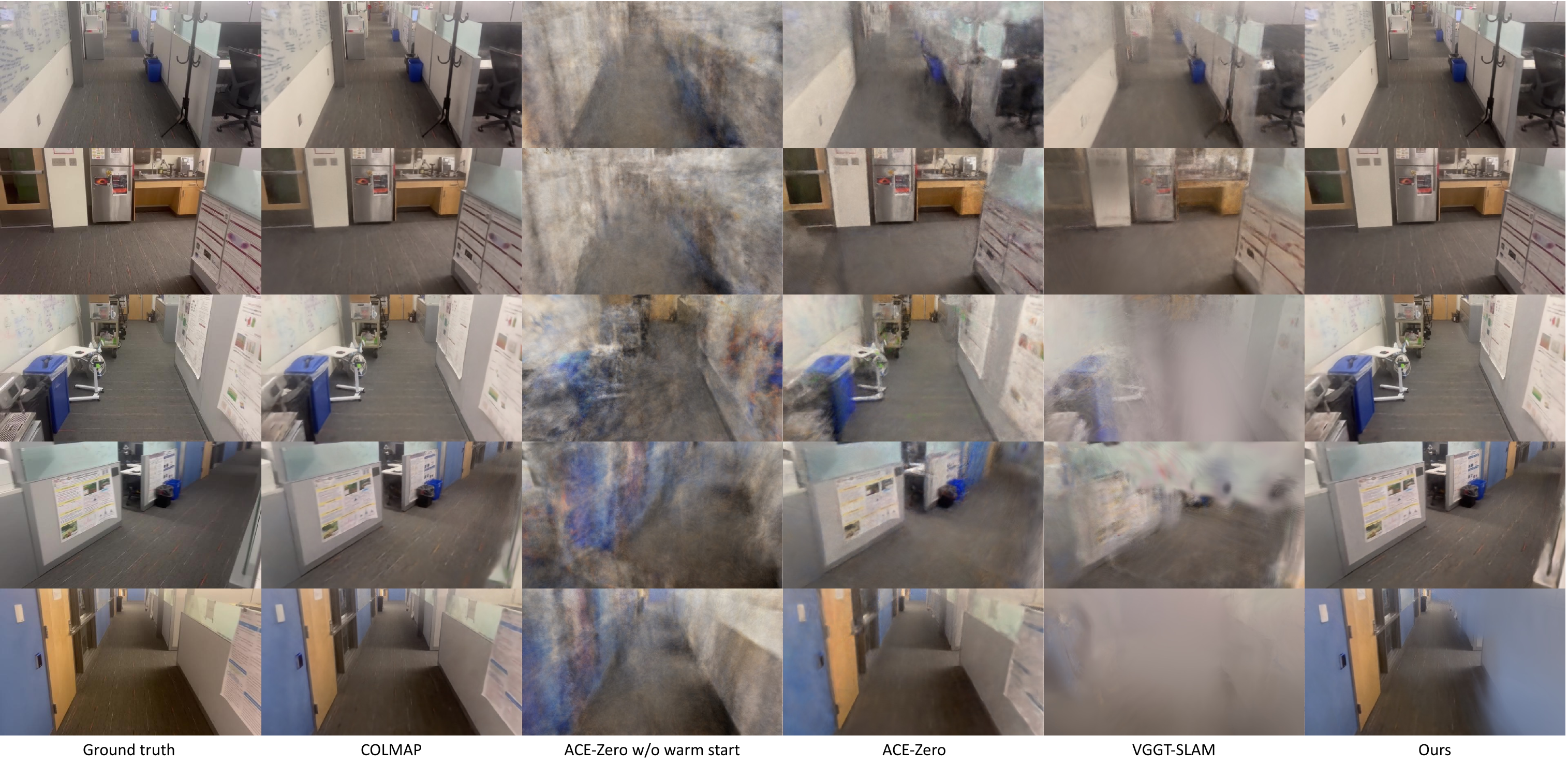}
\caption{Qualitative comparison from the \textit{office loop} video that were used for the teaser example. Although ACE-Zero, even when initialized with COLMAP-generated camera poses, tends to exhibit artifacts and blurriness in synthesized novel views, our proposed method consistently delivers high-fidelity renderings. This improvement reflects a more globally consistent and structurally coherent scene reconstruction.
}
\label{sup:fig:office_loop_nerf}
\end{figure*}

\paragraph{Details of the teaser example.} For the teaser figure, we use the the example video \textit{office loop} from \cite{maggio2025vggt} which consists of 473 frames capturing movement around an office environment. We provide the full qualitative comparison in Fig.~\ref{sup:fig:teaser} with sample image frames in Fig.~\ref{sup:fig:office_loop_frames}. As illustrated in Fig.~\ref{sup:fig:office_loop_frames}, ACE-Zero exhibits suboptimal performance, achieving PSNR values of 19.1 and 15.6 with and without COLMAP initialization, respectively. This highlights that the vanilla ACE-Zero configuration can incorrectly refine camera poses and fails to recover, even with additional iterations and accurate initialization from COLMAP. Furthermore, VGGT-SLAM, despite correctly identifying loop closures, does not recover sufficiently accurate camera poses, yielding a PSNR of 17.1. In contrast, our method achieves a PSNR of 25.6, which is comparable to COLMAP’s PSNR of 28.1. In addition, we provide qualitative comparisons in Fig.~\ref{sup:fig:office_loop_nerf}.

\paragraph{Full experimental results.} We present both SSIM and LPIPS scores for 7-Scenes and Mip-NeRF 360 datasets in Table~\ref{sup:tab:sevenscenes_SSIM} and \ref{sup:tab:mip_SSIM}, respectively. We also present SSIM and LPIPS scores for Tanks and Temples dataset in Table~\ref{sup:tab:t2_SSIM} and \ref{sup:tab:t2_LPIPS}, respectively. We observe that both SSIM and LPIPS scores behave similarly to the PSNR results presented in the main script.
\begin{table}[t]
\bgroup
    \centering%
    \large
    \adjustbox{valign=t,width=\linewidth}{%
        \begin{tabular}{cc||cc||cccccccc}
             & & & & & & & & & & &   \\ 
            \toprule
            \multicolumn{2}{c||}{~} &%
            \multicolumn{2}{c||}{~Pseudo Ground Truth~} &%
            \multicolumn{8}{c}{~All Frames~} 
             \\
            \hhline{~~----------}
             ~ &%
            \hspace{-6pt}\parbox[t]{0mm}{\multirow{2}{*}{\rotatebox[origin=c]{90}{Frames\hspace{-12pt}}}}
            &%
            ~COLMAP~ &%
            ~COLMAP~ &%
            \multicolumn{2}{c}{~ACE0~} &%
            \multicolumn{2}{c}{~ACE0~} &%
            \multicolumn{2}{c}{~Ours~} &%
            \multicolumn{2}{c}{~Ours~} 
             \\
             &%
             ~ &%
            SSIM ($\uparrow$) &%
            LPIPS ($\downarrow$) &%
            \multicolumn{2}{c}{SSIM ($\uparrow$)} &%
            \multicolumn{2}{c}{LPIPS ($\downarrow$)} &%
            \multicolumn{2}{c}{SSIM ($\uparrow$)} &%
            \multicolumn{2}{c}{LPIPS ($\downarrow$)} 
             \\
            \midrule 
             Chess & 6k & {\cellcolor{TableDarkGreen}0.86} & {\cellcolor{TableDarkGreen}0.17} & \multicolumn{2}{c}{\begin{tabular}{cc} & ~{0.84}~ \end{tabular}} &  \multicolumn{2}{c}{\begin{tabular}{cc} & ~{0.19}~ \end{tabular}} &  \multicolumn{2}{c}{\begin{tabular}{cc} & ~{0.85}~  \end{tabular}} &  \multicolumn{2}{c}{\begin{tabular}{cc} & ~{0.18}~ \end{tabular}}  \\
             & & & & & & & & & & &  \\[-10pt]
 Fire & 4k & {\cellcolor{TableDarkGreen}0.73} & {0.27}  & \multicolumn{2}{c}{\begin{tabular}{cc} & ~{0.68}~  \end{tabular}} &  \multicolumn{2}{c}{\begin{tabular}{cc} & ~{0.30}~ \end{tabular}} &  \multicolumn{2}{c}{\begin{tabular}{cc} & ~{\cellcolor{TableDarkGreen}0.73}~ \end{tabular}} &  \multicolumn{2}{c}{\begin{tabular}{cc} & ~{\cellcolor{TableDarkGreen}0.26}~ \end{tabular}}  \\
             & & & & & & & & & & & \\[-10pt]
 Heads & 2k & 0.82 & {0.26}  & \multicolumn{2}{c}{\begin{tabular}{cc} & ~{0.80}~ \end{tabular}} &  \multicolumn{2}{c}{\begin{tabular}{cc} & ~{0.28}~ \end{tabular}} &  \multicolumn{2}{c}{\begin{tabular}{cc} & ~{\cellcolor{TableDarkGreen}0.84}~ \end{tabular}} &  \multicolumn{2}{c}{\begin{tabular}{cc} & ~{\cellcolor{TableDarkGreen}0.25}~ \end{tabular}}  \\
             & & & & & & & & & & &  \\[-10pt]
 Office & 10k & 0.85 & {0.25}  & \multicolumn{2}{c}{\begin{tabular}{cc} & ~{0.82}~ \end{tabular}} &  \multicolumn{2}{c}{\begin{tabular}{cc} & {0.27} \end{tabular}} &  \multicolumn{2}{c}{\begin{tabular}{cc} & ~{\cellcolor{TableDarkGreen}0.86}~ \end{tabular}} &  \multicolumn{2}{c}{\begin{tabular}{cc} & ~{\cellcolor{TableDarkGreen}0.24}~ \end{tabular}}  \\
             & & & & & & & & & & &  \\[-10pt]
 Pumpkin & 6k & {0.84} & 0.23 & \multicolumn{2}{c}{\begin{tabular}{cc} & ~{0.83}~ \end{tabular}} &  \multicolumn{2}{c}{\begin{tabular}{cc} & ~{0.24}~ \end{tabular}} &  \multicolumn{2}{c}{\begin{tabular}{cc} & ~{\cellcolor{TableDarkGreen}0.87}~ \end{tabular}} &  \multicolumn{2}{c}{\begin{tabular}{cc} & ~{\cellcolor{TableDarkGreen}0.22}~ \end{tabular}}  \\
             & & & & & & & & & & & \\[-10pt]
 RedKitchen & 12k & {\cellcolor{TableDarkGreen}0.80} & {\cellcolor{TableDarkGreen}0.25}  & \multicolumn{2}{c}{\begin{tabular}{cc} & ~{0.77}~ \end{tabular}} &  \multicolumn{2}{c}{\begin{tabular}{cc} & ~{0.29}~ \end{tabular}} &  \multicolumn{2}{c}{\begin{tabular}{cc} & ~{0.78}~ \end{tabular}} &  \multicolumn{2}{c}{\begin{tabular}{cc} & ~{0.27}~ \end{tabular}}  \\
             & & & & & & & & & & &  \\[-10pt]
 Stairs & 3k & 0.59 & {0.50} & \multicolumn{2}{c}{\begin{tabular}{cc} & ~{0.62}~ \end{tabular}} &  \multicolumn{2}{c}{\begin{tabular}{cc} & ~{0.45}~  \end{tabular}} & \multicolumn{2}{c}{\begin{tabular}{cc} & ~{\cellcolor{TableDarkGreen}0.76}~  \end{tabular}} &  \multicolumn{2}{c}{\begin{tabular}{cc} & ~{\cellcolor{TableDarkGreen}0.29}~  \end{tabular}}  \\
             & & & & & & & & & & & \\[-10pt]
\hhline{~-----------}
             & & & & & & & & & & &  \\[-10pt]
             Average &      & 0.78 & {0.28} & \multicolumn{2}{c}{\begin{tabular}{cc} & ~{0.76}~  \end{tabular}} &  \multicolumn{2}{c}{\begin{tabular}{cc} & {0.29}  \end{tabular}} &  \multicolumn{2}{c}{\begin{tabular}{cc} & ~{\cellcolor{TableDarkGreen}0.81}~  \end{tabular}} &  \multicolumn{2}{c}{\begin{tabular}{cc} & ~{\cellcolor{TableDarkGreen}0.24}~  \end{tabular}}  \\
             & & & & & & & & & & & \\[-10pt]
\bottomrule
      \end{tabular}
    }
    \caption{
        Quantitative comparisons on 7-Scenes dataset.
        We report the pose accuracy via view synthesis as \textbf{SSIM} ($\uparrow$) and \textbf{LPIPS} ($\downarrow$). \TCBTableDarkGreen{Dark green} indicate the best results. We report results for COLMAP (default).
    }%
    \label{sup:tab:sevenscenes_SSIM}%
%
\egroup
\end{table}

\begin{table}
\bgroup
    \centering%
    \large
    \adjustbox{valign=t,width=\linewidth}{%

    }
    \caption{
        Quantitative comparisons on Mip-NeRF 360 dataset.
        We report the pose accuracy via view synthesis as \textbf{SSIM} ($\uparrow$) and \textbf{LPIPS} ($\downarrow$). \TCBTableDarkGreen{Dark green} indicate the best results. We report results for COLMAP (default). 
    }%
    \label{sup:tab:mip_SSIM}%
%
\egroup
\end{table}

\begin{table}
    \centering%
    \setlength{\tabcolsep}{1pt}%
    \footnotesize%
    \adjustbox{valign=t,width=1\linewidth}{%
    %
    }
    \caption{Quantitative comparisons on Tanks and Temples dataset. We report the pose accuracy via view synthesis as \textbf{SSIM} ($\uparrow$). \TCBTableDarkGreen{Dark green} indicate the best results, \textbf{excluding} COLMAP.
    }%
    \label{sup:tab:t2_SSIM}%
\end{table}
\begin{table}
    \centering%
    \setlength{\tabcolsep}{1pt}%
    \footnotesize%
    \adjustbox{valign=t,width=1\linewidth}{%
    %
} 
         \\
        \bottomrule
    \end{tabular}
    }
    \caption{Quantitative comparisons on Tanks and Temples dataset. We report the pose accuracy via view synthesis as \textbf{LPIPS} ($\downarrow$). \TCBTableDarkGreen{Dark green} indicate the best results, \textbf{excluding} COLMAP.
    }%
    \label{sup:tab:t2_LPIPS}%
\end{table}

\paragraph{Reolcalization performance on 7-Scenes (Table~\ref{sup:tab:reloc}).} We compare with the supervised relocalizer ACE~\cite{brachmann2023accelerated} by training with both KinectFusion and COLMAP mapping poses. As pointed out in \cite{brachmann2024acezero}, ACE achieves almost perfect relocalization rate under the 5cm, 5$^{\circ}$ error threshold as the evaluation is performed against COLMAP query poses. We show that our approach can further improve the relocalization performance by providing a more globally aligned poses.


\begin{table}[t]
\bgroup
    \centering%
    \large
    \adjustbox{valign=t,width=\linewidth}{%
        \begin{tabular}{ccccc}
             & & & &     \\ 
            \toprule
             ~ &%
            
            ~ACE~\cite{brachmann2023accelerated}~ &%
            ~ACE~\cite{brachmann2023accelerated}~ &%
            {~ACE0\cite{brachmann2024acezero}~} &%
            {~Ours~} 
             \\
             Supervision&%
             KinextFusion &
            COLMAP &%
            self-supervised &%
            self-supervised 
             \\
            \midrule 
             Chess  & 96.0\% & 100.0\% & 100.0\% &   ~100.0\%~    \\
             & & & &    \\[-10pt]
 Fire & 98.4\% & 99.5\%  & 98.8\% &   ~99.0\%~   \\
              & & & &   \\[-10pt]
 Heads  & 100.0\% & 100.0\% &  100.0\%&  100.0\%     \\
              & & & &    \\[-10pt]
 Office & 36.9\% & 100.0\%  &  99.1\%  &   99.1\%     \\
              & & & &    \\[-10pt]
 Pumpkin & 47.3\% & 100.0\% &  99.9\%  &   100.0\%    \\
              & & & &   \\[-10pt]
 Redkitchen  & 47.8\% & 98.9\%  &  98.1\%  &   98.5\%     \\
              & & & &    \\[-10pt]
 Stairs & 74.1\%& 85.0\% &  61.0\%  &  75.1\%      \\
              & & & &  \\[-10pt]
\hline
             &  & & &   \\[-10pt]
             Average &      71.5\% & 97.6\% &  93.8\%   &   96.0\%    \ \\
              & & & &   \\[-10pt]
\bottomrule
      \end{tabular}
    }
    \caption{
        Relocalization performance comparisons on 7-Scenes dataset. We follow \cite{brachmann2024acezero} and report percentage below 5cm, 5$^{\circ}$ error which are computed with respect to COLMAP pseudo ground truth.}%
    \label{sup:tab:reloc}%
%
\egroup
\end{table}

\paragraph{Uncertainty metrics weight sensitivity (Table~\ref{sup:tab:alpha_sensitivity}).}
We evaluate the impact of varying the weighting parameters $\alpha_1$ and $\alpha_2$ in Eq.~(9), which combine epipolar and optical flow errors into the confidence score. Our best-performing configuration is $\alpha_1 = 0.4$ and $\alpha_2 = 0.6$, achieving \textbf{21.7 dB} PSNR on the 7-Scenes dataset and \textbf{24.3 dB} PSNR on the Mip-NeRF 360 dataset. Table~\ref{sup:tab:alpha_sensitivity} reports performance for different weight ratios. We observe that moderate variations in $\alpha$ (e.g., 0.3:0.7 or 0.5:0.5) have minimal impact on performance, indicating robustness to weight changes. However, extreme imbalance significantly degrades accuracy. When epipolar weighting dominates (e.g., $\alpha_1 \geq 0.7$), performance drops due to sensitivity to pose initialization errors and lack of local consistency. Conversely, optical flow only configurations ($\alpha_2 = 1.0$) improve robustness in indoor scenes but fail to enforce global consistency, leading to drift in large-scale or mixed environments. The best trade-off is achieved with $\alpha_1 = 0.4$ and $\alpha_2 = 0.6$, which balances global and local geometric cues effectively across both datasets. This suggests that optical flow provides strong local evidence for texture-rich indoor scenes, while epipolar constraints remain essential for maintaining global structure in complex environments.
\begin{table}[h]
\centering
\begin{tabular}{c|c|c}
\hline
$\alpha_1 : \alpha_2$ & 7-Scenes & Mip-NeRF 360 \\
\hline
0.4 : 0.6 (Best) & \textbf{21.7} & \textbf{24.3} \\
0.5 : 0.5       & 21.5 & 24.1 \\
0.3 : 0.7       & 21.6 & 24.2 \\
0.6 : 0.4       & 21.3 & 23.8 \\
0.7 : 0.3       & 21.0 & 23.5 \\
1.0 : 0.0       & 20.5 & 22.8 \\
0.0 : 1.0       & 21.2 & 23.6 \\
\hline
\end{tabular}
\caption{Ablations on different $\alpha$ ratios on 7-Scenes and Mip-NeRF 360 datasets. We report the pose accuracy via view synthesis as PSNR in dB.}
\label{sup:tab:alpha_sensitivity}
\end{table}

\paragraph{Full runtime breakdown.} For a 200 images frames dataset, our additional pose graph optimization step requires approximately 30 seconds per reconstruction iteration, compared to 150 seconds for vanilla ACE-Zero. Of this 30-second overhead, the actual optimization accounts for only 0.016 seconds, while the remaining time is spent on graph construction and uncertainty metric computation.

The scalability of this step depends primarily on graph construction rather than optimization. The optimization itself is highly efficient, as it operates on a sparse graph and optimizes only camera poses, resulting in near-linear complexity with respect to the number of nodes and edges. In contrast, graph construction involves pairwise consistency checks and uncertainty metric computation, which can scale quadratically with the number of images if all pairs are considered. This explains why graph construction dominates the runtime: it requires dense correspondence verification and optical flow estimation across multiple image pairs, whereas the optimization step leverages sparse factor graphs and converges rapidly. For larger datasets, pruning strategies or retrieval-based filtering will be essential to maintain scalability.

\paragraph{Effect of removing optical flow or epipolar metric individually (Table~\ref{sup:tab:uncertainty_metrics}).} 
We analyze the effect of incorporating uncertainty-aware metrics into pose graph optimization. Starting from vanilla PGO with graph construction, we progressively add epipolar and optical flow metrics. The full configuration uses both metrics. Table~\ref{sup:tab:uncertainty_metrics} summarizes the results for the 7-Scenes and Mip-NeRF 360 datasets. We observe that adding uncertainty-aware metrics significantly improves pose graph optimization over vanilla PGO. Epipolar constraints provide global geometric consistency, yielding a 1.0 dB improvement on both datasets. Optical flow offers dense local correspondences, which are particularly beneficial in texture-rich indoor scenes, leading to 1.6 dB improvement over vanilla PGO. Combining both metrics achieves the highest gains (2.3 dB and 2.2 dB): epipolar constraints enforce global structure, while optical flow enhances local robustness. This complementary effect highlights the importance of integrating multiple geometric priors for reliable pose refinement.

\begin{table}[h]
\centering
\begin{tabular}{l|c|c}
\hline
Configuration & 7-Scenes & Mip 360 \\
\hline
Vanilla PGO (with construction) & 18.5 & 21.0 \\
+ Epipolar metric                & 19.5 & 22.0 \\
+ Optical flow metric & 20.1 & 22.3 \\
+ Both metrics                   & \textbf{20.8} & \textbf{23.2} \\
\hline
\end{tabular}
\caption{Ablations on different uncertainty-aware configurations. We report the pose accuracy via view synthesis as PSNR in dB.}
\label{sup:tab:uncertainty_metrics}
\end{table}

\begin{figure}[h]
\centering
\includegraphics[width=1\linewidth]{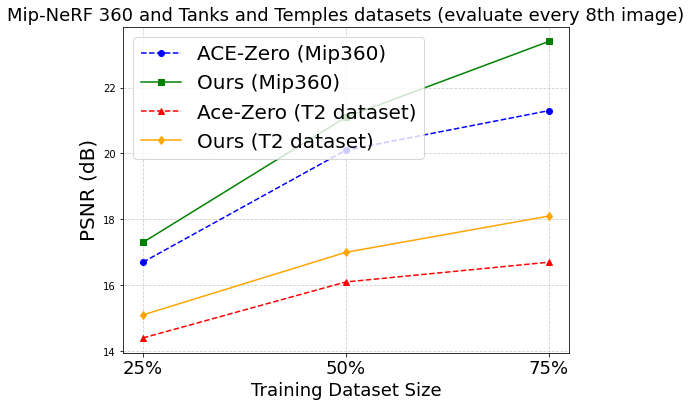}
\caption{Ablations on sparse training data for both Mip-NeRF 360 and Tanks and Temples (T2) datasets. We show that our method consistently demonstrates better robustness to limited training data.
}
\label{sup:fig:sparse_dataset}
\end{figure}

\paragraph{Full impact of spare data (Fig.~\ref{sup:fig:sparse_dataset}).} After including the results from the Tanks and Temples dataset, we identified an error in the previously generated figure. The figure has now been corrected, and the version in the main text should be disregarded. The updated figure presented here demonstrates that our method consistently exhibits superior robustness under limited training data conditions.

\section{Additional qualitative examples} \label{sup:qualitative}
\begin{figure*}
\centering
\includegraphics[width=1\linewidth]{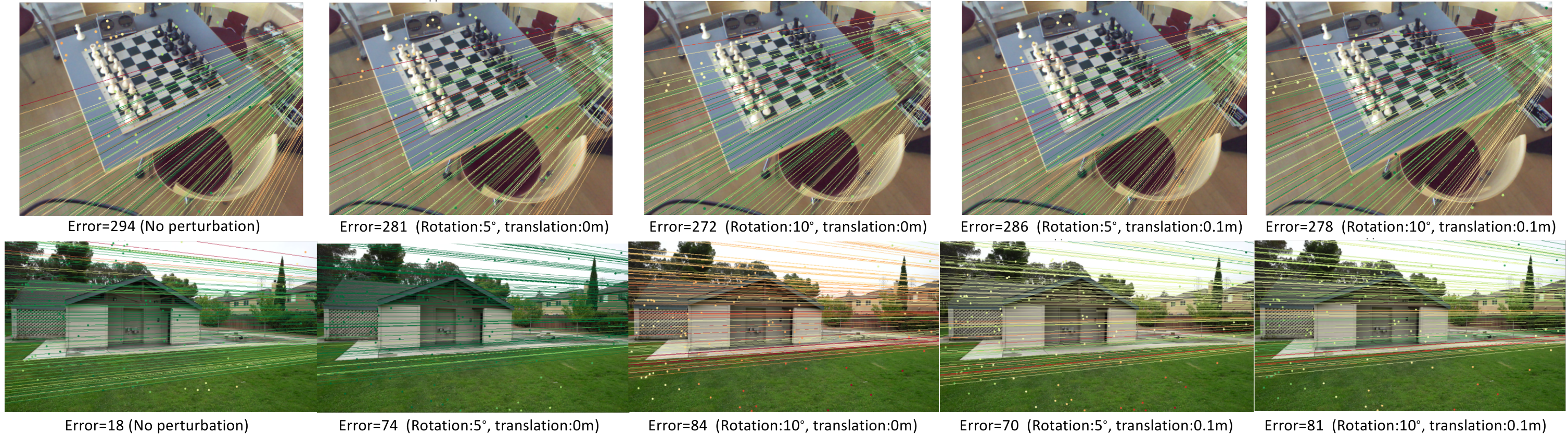}
\caption{Qualitative example of the epipolar-based uncertainty metric (top row: 7-Scenes dataset; bottom row: Tanks and Temples dataset). We overlay reprojected points with their corresponding epipolar lines, color-coded by error (green = low, red = high), to indicate geometric alignment quality under perturbed poses. In the top row, the highest error occurs when no perturbation is applied, suggesting the original camera pose is inaccurate; introducing rotation perturbations reduces the error. In the bottom row, from left to right, we observe a similar trend as in Fig. 2 of the main text: errors remain low without perturbation, increase with pose perturbations, and occasionally decrease when translation perturbations are added. This behavior highlights that epipolar error is sensitive to interactions between pose components and does not always correlate directly with pose accuracy. Consequently, we treat it as a heuristic confidence indicator rather than a strict geometric validator.
}
\label{sup:fig:epipolar}
\end{figure*}

\paragraph{Epipolar-based uncertainty metric (Fig.~\ref{sup:fig:epipolar}).}
To further analyze the behavior of the epipolar-based uncertainty metric, we apply synthetic perturbations to estimated camera poses. Specifically, we introduced rotational perturbations around the z-axis and translational perturbations along the z-axis to simulate varying degrees of pose deviation. For each pose, we randomly sampled 100 predicted 3D scene coordinates and reprojected them into the paired image. We then drew the corresponding epipolar lines and discarded any second-view projections that fell outside the image boundaries to ensure valid comparisons. Fig.~\ref{sup:fig:epipolar} illustrates qualitative results on two datasets (7-Scenes and Tanks and Temples), where reprojected points and their epipolar lines are color-coded by error magnitude (green = low, red = high). Interestingly, in some cases, the highest epipolar error occurs when no perturbation is applied, suggesting that the original pose estimate is inaccurate. Introducing small rotational perturbations can reduce this error, while translation perturbations occasionally lead to similar effects. These observations indicate that epipolar error is sensitive to interactions between pose components and does not consistently correlate with absolute pose accuracy. Consequently, we interpret this metric as a heuristic confidence indicator rather than a strict geometric validator.

\begin{figure*}
\centering
\includegraphics[width=1\linewidth]{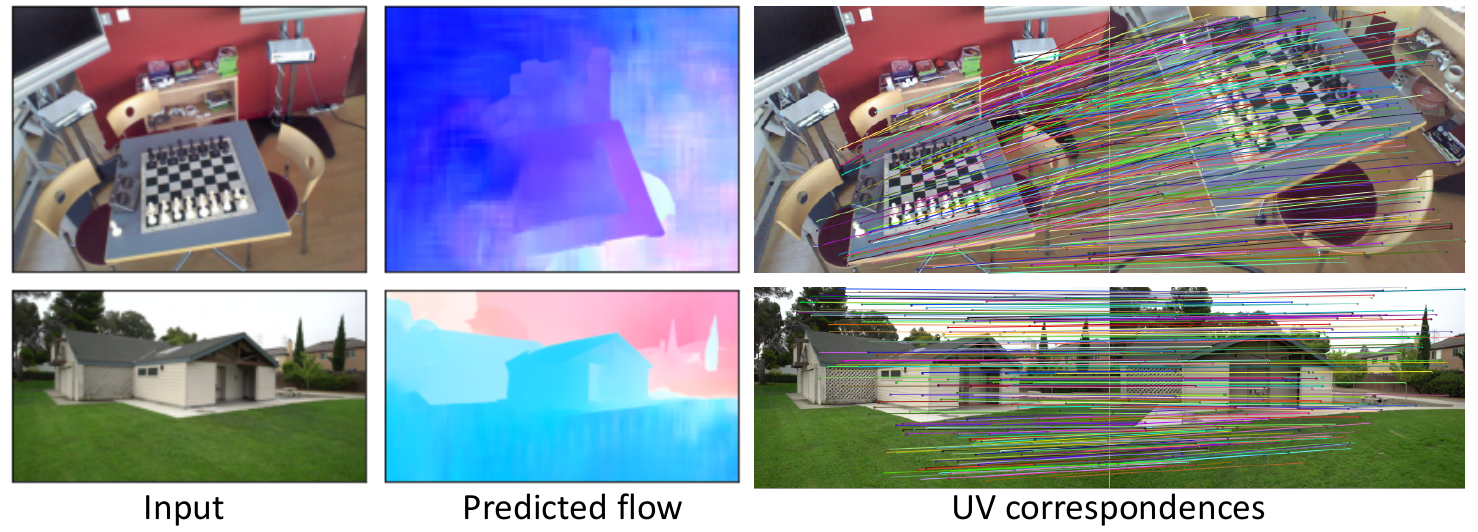}
\caption{Qualitative examples of optical flow predictions are shown (top row: 7-Scenes dataset; bottom row: Tanks and Temples dataset). From left to right: (i) an image from the input pair, (ii) predicted flow visualization where similar colors indicate similar motion directions, and (iii) UV correspondences overlaid on the image pair, illustrating that optical flow provides robust pixel-level matches across views.
}
\label{sup:fig:optical_exp}
\end{figure*}

\paragraph{Optical flow-based uncertainty metric.} We provide additional qualitative examples in Fig.~\ref{sup:fig:optical_exp}.

\begin{figure*}[h]
\centering
\includegraphics[width=1\linewidth]{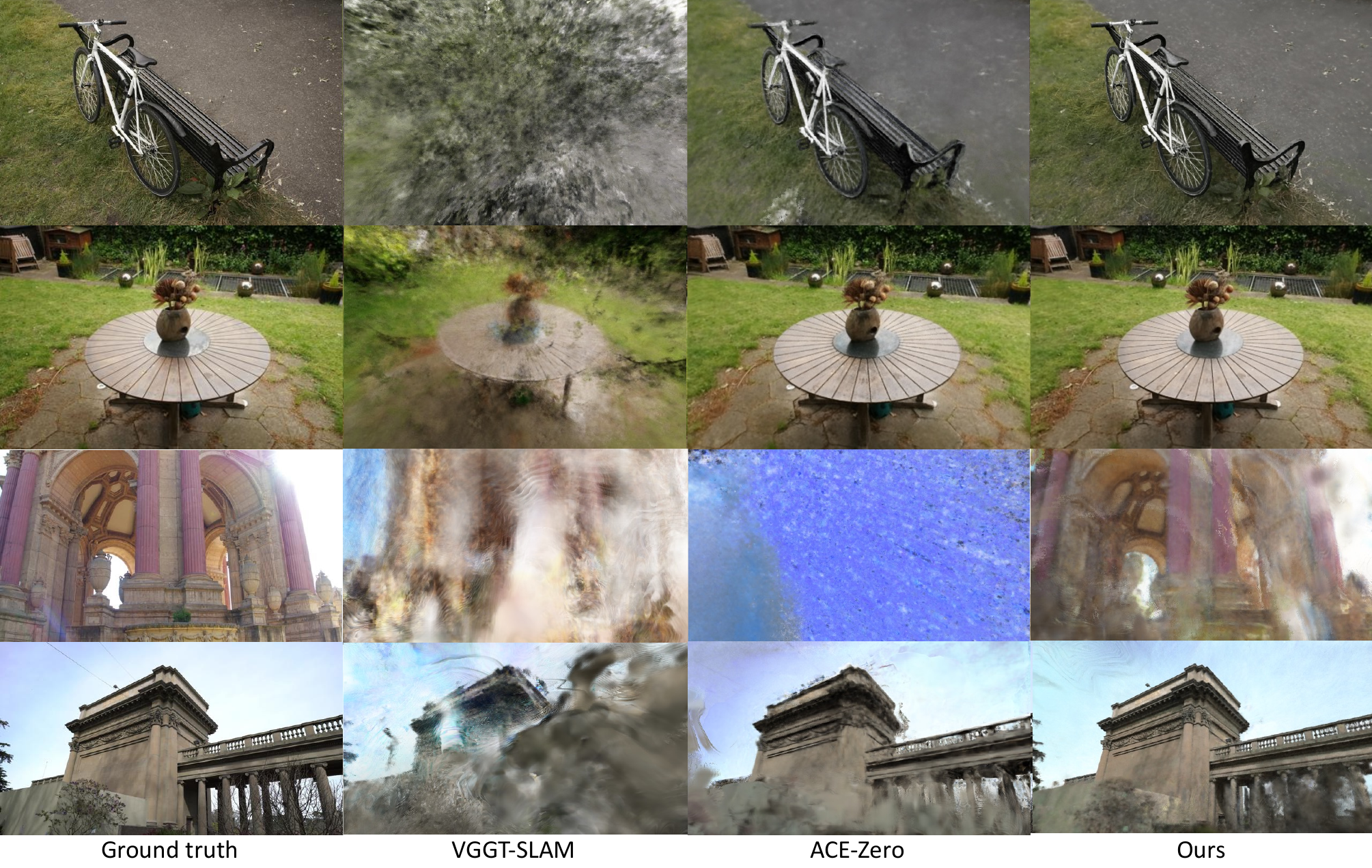}
\caption{Qualitative comparison on Mip-NeRF 360 (top two rows) and Tanks and Temples datasets (bottom two rows) with VGGT-SLAM~\cite{maggio2025vggt} and ACE-Zero~\cite{brachmann2024acezero}. While ACE-Zero performs well on small-scale scenes such as Mip-NeRF 360, it struggles with large, complex environments like Tanks and Temples, where our method maintains high rendering quality across diverse scenes.
}
\label{sup:fig:compairson}
\end{figure*}

\paragraph{Additional qualitative examples.} We provide additional qualitative examples in Fig.~\ref{sup:fig:compairson}.

\end{document}